\documentclass[english]{article}
\usepackage[latin9]{inputenc}
\usepackage{geometry}
\usepackage{float}
\usepackage{amsmath}
\usepackage{amssymb}
\usepackage{graphicx}

\makeatletter

\providecommand{\tabularnewline}{\\}
\newcommand{\lyxdot}{.}

\usepackage{times}
\date{}
\usepackage{babel}
\usepackage{epstopdf}
\newcommand{\q}{\quad\quad}

\makeatother

\usepackage{babel}
\begin{document}

\title{Learning Structured Outputs from Partial Labels using Forest Ensemble}

\author{Truyen Tran, Dinh Phung, Svetha Venkatesh\\
 Centre for Pattern Recognition and Data Analytics\\
 Deakin University, Australia}
\maketitle
\begin{abstract}
Learning structured outputs with general structures is computationally
challenging, except for tree-structured models. Thus we propose an
efficient boosting-based algorithm AdaBoost.MRF for this task. The
idea is based on the realization that a graph is a superimposition of trees. Different
from most existing work, our algorithm can handle \emph{partial labelling},
and thus is particularly attractive in practice where reliable labels
are often sparsely observed. In addition, our method works exclusively
on trees and thus is guaranteed to converge. We apply the AdaBoost.MRF
algorithm to an indoor video surveillance scenario, where activities
are modelled at multiple levels. 
\end{abstract}
\global\long\def\HCRF{HCRF}
 \global\long\def\HCRFs{HCRFs}
 \global\long\def\DCRF{DCRF}
 \global\long\def\DCRFs{DCRFs}
 \global\long\def\graph{\mathcal{G}}

\global\long\def\vertex{\mathcal{V}}
 \global\long\def\edges{\mathcal{E}}
 \global\long\def\C{\mathcal{C}}
 \global\long\def\clix{{\bf r}}
 \global\long\def\T{\mathcal{T}}
 \global\long\def\Loss{\mathcal{L}}
 \global\long\def\f{\mathbf{f}}
 \global\long\def\F{\mathbf{F}}
 \global\long\def\w{\mathbf{w}}

\global\long\def\E{\mathbb{E}}
 \global\long\def\Free{\mathcal{H}}
 \global\long\def\BigO{\mathcal{O}}
 \global\long\def\N{\mathcal{N}}
 \global\long\def\Data{\mathcal{D}}
 \global\long\def\U{\mathcal{U}}
 \global\long\def\I{\mathcal{I}}
 \global\long\def\M{\mathcal{M}}
 \global\long\def\obs{\mathbf{o}}

\global\long\def\states{\mathbf{x}}
 \global\long\def\visibles{\mathbf{v}}
 \global\long\def\hiddens{\mathbf{h}}
 \global\long\def\q{\quad}
 \global\long\def\LogZ{\mathcal{A}}

\section{Introduction\label{sec:intro}}

There has been a growing research interest in developing probabilistic
temporal graphical models for recognising human activities from sensory
data. In this paper we address an important aspect of the problem
in that there are multiple levels of abstraction, that is, an activity
is often composed of several sub-activities. A popular approach to
deal with such a hierarchical nature is to build a cascaded model:
each level is modelled separately, and the output of the lower levels
is subsequently used as the input for the upper levels \cite{oliver2004lrl}.
This approach is sub-optimal because the information at the higher
level is often very discriminative to infer about the lower levels,
but it is not modelled. Moreover, the layered approach often suffers
from the so-called \emph{cascading error} problem, as the error introduced
from the lower level will propagate to higher tasks.

A better and more holistic approach is to build a joint representation
at all layers. Emerging methods include generative/directed models
such as abstract hidden Markov models (AHMMs) \cite{Bui-et-al02},
hierarchical HMMs \cite{Nam-et-alCVPR05}, dynamic Bayesian networks
\cite{Gong-Xiang-ICCV03}, and their discriminative/undirected counterparts
such as hierarchical conditional random field (HCRF) \cite{liao2007epa},
and dynamic CRF (DCRF) \cite{Sutton-et-alJMLR07}. In general, generative
models are useful in jointly modelling the activity semantics and
observed sensory data, and thus dealing well with hidden variables.
On the contrary, discriminative models make less assumption about
the nature of the sensory data, and thus can potentially achieve higher
classification accuracy when the generative assumption is violated.

Our application of interest is the problem of activity recognition
in an indoor environment. The solution to this problem will provide
important technology for building intelligent surveillance systems.
To this end, we propose in this paper the use of the DCRF for modelling
activities at multiple scales. Such a problem is known to be difficult
due to the hierarchical nature of the activities. The DCRF used in
this paper is a generalisation of a powerful probabilistic formulation
known as conditional random field (CRF) \cite{lafferty01conditional}.
It is an expressive representation scheme that seamlessly integrates
domain knowledge, temporal regularities, and at the same time encodes
complex interdependency between semantic levels. However, its expressiveness
does come with great challenges, one of which is problem of parameter
estimation under arbitrary structures. It is known that parameter
estimation is generally intractable to perform exactly in a maximum
likelihood setting for general MRFs, except for tree structures.

One of the earliest and most popular methods to deal with this intractability
is to optimise the pseudo-likelihood instead \cite{Besag-75}, which
achieves its efficiency through limiting to individual nodes in the
graph and theirs neighbours. However, the method cannot handle missing
variables, which unfortunately often happen in real situations. Sampling-based
methods such as MCMC are theoretically attractive, but they are often
impractical for extremely slow convergence. The state-of-the-art methods
often involves approximate inference algorithms such as Pearl's belief
propagation (BP) \cite{Pearl88} or a more recent method introduced
by Wainwright, Jaakkola and Willsky (WJW) \cite{Wainwright-et-alIEEEIT05}.
These methods are efficient but their biggest problem is that there
is no guarantee on the convergence. In addition, the WJW inference
has not been applied for learning in conditional MRFs, an issue that
we will also explored in this paper.

Our main contribution in this paper is the introduction of AdaBoosted
Markov Random Forests (AdaBoost.MRF), a learning method that operates
only on Markovian trees at each step, but can be proved to achieve
global optimum under some mild assumptions. Distinct from most existing
work on discriminative models including the work of \cite{lafferty01conditional,Sutton-et-alJMLR07},
our AdaBoost.MRF can also handle partial labels) -- an important enhancement
for real-world applications.

In essence, our AdaBoost.MRF is based on a ranking-based multiclass
boosting algorithm called AdaBoost.MR \cite{schapire99improved}.
At each round, the AdaBoost.MRF selects the best trained spanning
tree of the network based on the performance on weighted error. The
data is adaptively re-weighted to address more hard-to-classify instances.
Each selected spanning tree is weighted and combined to recover the
original graph. Finally, the parameters of the graph are the convex
combination of all selected tree parameters. Since our method works
exclusively on trees, inference is very efficient and convergence
is guaranteed. We also prove that under mild assumptions, the AdaBoost.MRF
reaches the unique optimum. Furthermore, since the AdaBoost.MRF considers
all the variables in the MRFs, the partial labels problem can also
be effectively handled.

We evaluate AdaBoost.MRF on the activity data obtained in a indoor
video monitoring scenario. We compare our AdaBoost.MRF with the maximum
likelihood method, which used BP and WJW as inference engines. To
evaluate the effectiveness of the discriminative DCRFs against generative
methods, we implement a variant of the layered hidden Markov models
(LHMMs)~\cite{oliver2004lrl} which can handle partially observed
state variables to make it comparable with the DCRFs considered in
this paper. We also show that the multi-level DCRFs perform better
than the flat-CRFs as more information is added.

The paper will continue with a related background in Section~\ref{sec:bgr}.
Section~\ref{sec:ada} introduces the AdaBoost.MRF algorithm and
discusses its convergence. Section~\ref{sec:DCRF-ML} shows how we
model and learn multi-level activities with DCRFs, followed by Section~\ref{sec:exp}
to present the experimental results. Discussion is given in Section~\ref{sec:discuss}
and conclusion in Section~\ref{sec:con}.

\section{Related Work and Background\label{sec:bgr}}

Since the starting point of our work is the conditional random field
and multiclass boosting, we shall provide a brief related work and
background to these two problems respectively in this section.

\subsection{Conditional random fields\label{sec:bgr-crf}}

Our work is based on conditional random fields (CRFs) \cite{Kumar-Hebert-ICCV03,lafferty01conditional,Sminchisescu-et-alICCV05,Truyen:2007}.
A Markov random field (e.g., see Figure \ref{fig:spanning}) is an
undirected graph $\graph=(\vertex,\edges)$ which represents a joint
distribution of state variables $\states=(x_{1},x_{2},...,x_{n})\in\mathcal{X}$,
where each $x_{r}$ corresponds to a node $r\in\mathcal{V}$ in the
graph. For simplicity we assume these variables have the same domain
and receive assignments from a fixed set of discrete values $S=\{1,2,...,|S|\}$.
In conditional MRF, the graph is further associated with a data observation
$\obs$ in which no knowledge of the structure within the observation
is required. The edge-set $\edges$ of the graph specifies a set of
cliques $C$, each of which supports a feature vector $\f(x_{c},\obs)$
that maps the observation and the clique-based state variable $x_{c}$.
Let $\F(\states,\obs)=\sum_{c\in C}\f(x_{c},\obs)$ be the global
feature vector. A standard CRF defines a conditional distribution
of the state variable given the observation $\obs$ in an exponential
family distribution as: 
\begin{eqnarray}
P(\states|\obs;\w) & = & \exp\{\langle\w,\F(\states,\obs)\rangle-\LogZ(\obs)\}\label{expo-dis}
\end{eqnarray}
where $\w$ is the feature weight vector, $\left\langle \cdot,\cdot\right\rangle $
is the inner product, and $\LogZ(\obs)$ is the log-partition function,
i.e., $\LogZ(\obs)=\log\sum_{\states\in\mathcal{X}}\exp\langle\w,\F(\states,\obs)\rangle$.

Parameter learning in CRFs is often based on the maximum likelihood
criterion. Given a training set $\Data=\{\states^{(i)},\obs^{(i)}\}_{i=1}^{D}$,
the goal is to find the maximiser of the (log) likelihood with respect
to $\w$: 
\begin{eqnarray}
\Loss(\w) & = & \frac{1}{D}\sum_{i=1}^{D}\log P(\states^{(i)}|\obs^{(i)};\w)\\
 & = & \frac{1}{D}\sum_{i=1}^{D}\left\{ \langle\w,\F(\states^{(i)},\obs^{(i)})\rangle-\LogZ(\obs^{(i)})\right\} 
\end{eqnarray}
This function is concave in $\w$ thus it has a unique global maximum.
Optimisation methods often require the gradient, which can obtained
as in a standard exponential family case as: 
\begin{eqnarray*}
\nabla\Loss(\w)=\frac{1}{D}\sum_{i=1}^{D}\sum_{c\in C}\left\{ \f(x_{c}^{(i)},\obs^{(i)})-\E_{x_{c}|\obs}[\f(x_{c},\obs^{(i)})]\right\} 
\end{eqnarray*}
where $\E_{x_{c}|\obs}[\cdot]$ is the expectation evaluated with
respect to $P(x_{c}|\obs)$. 

However, inference in general networks is known to be intractable
except for trees with limited tree-widths. For tree-structures, a
well-known method is a two-pass belief propagation procedure \cite{Pearl88}
that takes $\BigO(2|\vertex||S|^{2})$ time to compute all quantities
needed for learning, where $|\vertex|$ is number of nodes in the
tree. 

For structures other than trees extract inference is intractable and
approximate methods such as mean fields and loopy belief propagation
(BP) are widely used. A more recently proposed method by Wainwright,
Jaakkola and Willsky (WJW) \cite{Wainwright-et-alIEEEIT05} also offers
an interesting alternative to compute the so-called pseudo-marginals
based on minimising the upper bound of the log-partition function.
Like BP, the WJW is an efficient message passing scheme. However,
the BP is not guaranteed to converge and there has not been any formal
proof nor extensive empirical evaluation for WJW found in literature.
Our AdaBoost.MRF, in contrast, works on tree inference and thus is
guaranteed to converge with known analytical complexity. Besides,
to the best of our knowledge, we are the first to perform learning
using WJW in the conditional MRFs setting. CRFs have also been applied
to activity recognition recently \cite{liao2007epa,Truyen:2006,sminchisescu2006cmc,taycher2006crp,quattoni2007hcr,vail2007crf}.
These works have reported promising results. However, due to the inherent
intractability of the general structures, either simple chains and
trees have been assumed or approximate inference methods have been
used.

\subsection{Learning Structured Output with Partial Labels}

Conditional random fields are an example of structured output models
\cite{tsochantaridis2005large,taskar2005learning,daume2005learning}.
Learning in structured output models can be based on principles other
than maximum likelihood, for example, large-margin \cite{tsochantaridis2005large,taskar2005learning},
or search \cite{daume2005learning}. With the latter, the computation
of feature expectation is replaced by finding the most probable labelling. 

Learning with partial labels have been addressed in the past decade
\cite{Truyen:2005a,Truyen:2005b,Truyen:2006}. Related problems include
weak supervision \cite{vezhnevets2012weakly} and indirect supervision
\cite{chang2010structured}. Partial labels arise when only some components
of $\mathbf{x}$ are observed, i.e., $\states=\left\{ \visibles,\hiddens\right\} $
where $\mathbf{v}$ and $\mathbf{h}$ are observed and missing components
respectively. In the CRF setting, parameter learning requires to maximise
the conditional incomplete log-likelihood instead, which can be shown
to be:

\begin{align}
\mathcal{\Loss}(\w) & =\frac{1}{D}\sum_{i=1}^{D}\log P(\visibles^{(i)}|\obs^{(i)})\label{ll}\\
 & =\frac{1}{D}\sum_{i=1}^{D}\LogZ(\visibles^{(i)},\obs^{(i)})-\LogZ(\obs^{(i)})
\end{align}
where $\LogZ(\visibles,\obs)=\log\sum_{\hiddens}\exp\langle\w,\F(\visibles,\hiddens,\obs)\rangle$.
The gradient $\nabla\Loss(\w)$ can now be derived as:

\begin{align}
\nabla\Loss(\w) & =\frac{1}{D}\sum_{i=1}^{D}\sum_{c\in C}\left(\sum_{h_{c}}\E_{h_{c}|\mathbf{v},\obs}[\f(v_{c}^{(i)},h_{c},\obs^{(i)})]-\E_{x_{c}|\obs}[\f(x_{c},\obs^{(i)})]\right)\label{ll-grad}
\end{align}
where $h_{c}$ denotes the missing components associated with clique
$c$. Equations (\ref{ll}) and (\ref{ll-grad}) reveal that learning
depends on the inference ability to compute $\LogZ(\visibles,\obs)$,
$\LogZ(\obs)$ and local clique distributions $P(x_{c}|\obs)$.

\subsection{Ranking-based Multiclass Boosting \label{sec:bgr-boosting}}

This section reviews a multi-class boosting algorithm known as AdaBoost.MR~\cite{schapire99improved},
based on which our work will be developed. We adopt the functional
view of boosting from~\cite{Manson-et-al00} in this paper.

Given a pool of weak learners $\{h_{m}(\states,\obs)\}$, in a boosting
setting we seek to learn a subset $\{h_{k}(\states,\obs)\}_{k=1}^{K}$
and their corresponding weights $\{w_{k}\}_{k=1}^{K}$. Denote $H(\states,\obs)=\sum_{k=1}^{K}w_{k}h_{k}(\states,\obs)$
to be a final classifier that outputs the prediction: 
\begin{eqnarray}
\hat{\states}=\arg\max_{\states\in\mathcal{X}}H(\states,\obs).\label{max}
\end{eqnarray}
$H(\states,\obs)$ is also known as strong learner in the boosting
literature. For each training instance $(\states^{(i)},\obs^{(i)})$,
we would expect that $\states^{(i)}=\hat{\states}^{(i)}$, and thus
we expect: 
\begin{equation}
H(\states^{(i)},\obs^{(i)})\ge H(\states,\obs^{(i)})
\end{equation}
for all $\states\in\mathcal{X}$ and $i=1,2,...,D$. Whenever there
exists an observation $\obs$ that invalidates this assertion, the
system suffers a loss. The \emph{rank loss }is defined as 
\begin{eqnarray}
{\Loss}_{rank}=\frac{1}{D}\sum_{i=1}^{D}\sum_{\states}\mathbb{I}\left\{ H(\states,\obs^{(i)})-H(\states^{(i)},\obs^{(i)})>0\right\} \label{bgr:rank-loss}
\end{eqnarray}
where $\mathbb{I}\left\{ \cdot\right\} $ is the indicator function.
This rank loss is basically the number of possibilities where the
system misclassifies the data. The loss vanishes if the system correctly
classifies all the data instances.

However, the rank-loss in Equation~\ref{bgr:rank-loss} is difficult
to minimise. Therefore, we resort to the exponential-loss, which is
a smooth, convex upper-bound of the rank-loss: 
\begin{eqnarray}
{\Loss}_{exp}=\frac{1}{D}\sum_{i=1}^{D}\sum_{\states}\exp\{H(\states,\obs^{(i)})-H(\states^{(i)},\obs^{(i)})\}\label{bgr:exp-loss}
\end{eqnarray}
Term-by-term comparison of Equations~\ref{bgr:rank-loss} and Equation~\ref{bgr:exp-loss}
it can easily verify that ${\Loss}_{exp}$ is indeed the upper-bound
of ${\Loss}_{rank}$ up to a constant%
\footnote{To see the connection between the exponential loss and the log-likelihood,
assume a conditional distribution 
\begin{eqnarray}
P(\states|\obs)=\frac{1}{Z(\obs)}\exp\{H(\states,\obs)\}\label{BMN-exp-model}
\end{eqnarray}
where $Z(\obs)=\sum_{x}\exp\{H(\states,\obs)\}$ is the normalisation
constant. This assumption makes sense because the prediction is identical
to the Maximum A Posteriori: 
\begin{eqnarray}
\hat{\states}=\arg\max_{x}H(\states,\obs)=\arg\max_{x}P(\states|\obs)
\end{eqnarray}
Substituting $P(\states|\obs)$ into Equation~\ref{bgr:exp-loss}
yields 
\begin{eqnarray}
{\Loss}_{exp}(\w)=\frac{1}{D}\sum_{i=1}^{D}\frac{1}{P(\states^{(i)}|\obs^{(i)})}
\end{eqnarray}
This appears similar to the log-loss used in the maximum likelihood
estimation 
\begin{eqnarray}
{\Loss}_{log}(\w)=\frac{1}{D}\sum_{i=1}^{D}\log\frac{1}{P(\states^{(i)}|\obs^{(i)})}
\end{eqnarray}
The difference between the exponential loss and the log-loss is about
the numerical scale, because of the $\log$ function in the log-loss.
However, in \cite{nips02-LT09} the authors show that the two losses
give very close results given enough data. This paper suggests that
boosting can be regarded as an (approximate) alternative for the maximum
likelihood estimation (MLE). From another related angle, boosting-style
MLE algorithms are derived in \cite{Friedman-et-al00,collins2002lra}.%
}.

The learning process in boosting is iterative, in that at each step
$l$, we greedily seek for an update of the functional $H(.)$ that
best reduces the loss: 
\begin{eqnarray}
H_{l} & \leftarrow & H_{l-1}+\alpha_{l}h_{j}\qquad\mbox{ where}\label{bgr:boosting-update1}\\
(\alpha_{l},j) & = & \arg\min_{\alpha,k}{\Loss}_{exp}(H_{l-1}+\alpha h_{k})\label{bgr:boosting-update2}
\end{eqnarray}
The standard AdaBoost.MR addresses only simple classification, where
the data of interest does not have any structure. For structured data
such as CRFs, boosting has been applied in \cite{NIPS2005_704}, but
the algorithm relies on the BP for approximate inference and does
not address the missing variables. Similarly, work in \cite{AA63,Dietterich-et-alICML04}
is limited to tractable CRFs.

\section{Forest Ensemble Algorithm: AdaBoost.MRF \label{sec:ada}}

In this section we present a novel boosting algorithm for parameter
estimation of general conditional random fields, termed as AdaBoost.MRF.
We consider the general case where the state label $\states$ may
have a visible component $\visibles$ and a missing component $\hiddens$,
i.e. $\states=\left\{ \visibles,\hiddens\right\} $.

\subsection{Forest Ensemble}

\begin{figure}[h]
\begin{centering}
\includegraphics[width=0.85\linewidth]{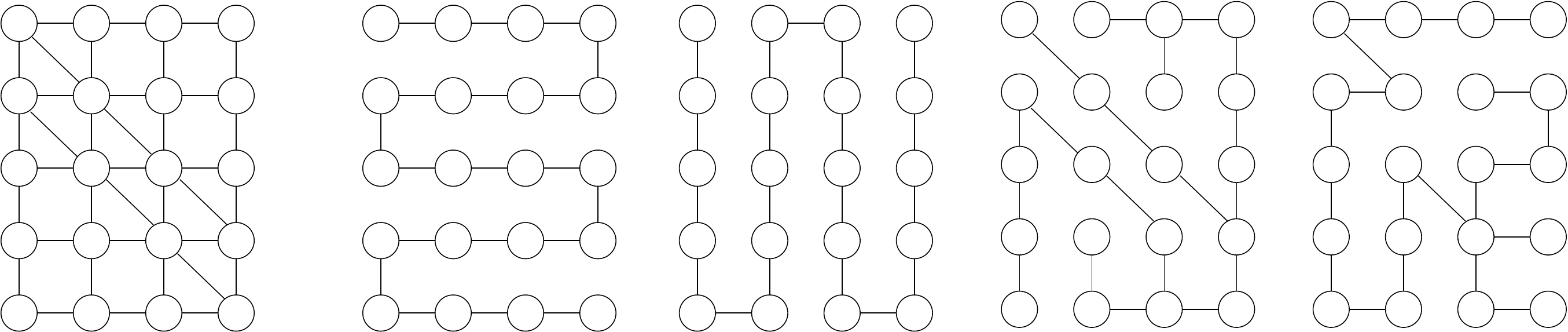} 
\par\end{centering}

\caption{An example of Markov network (left-most) and some spanning trees (right).\label{fig:spanning}}
\end{figure}

We formulate an objective function based on the exponential loss of
AdaBoost.MR. Recall from Section~\ref{sec:bgr-boosting} that given
training data pairs $(\states^{(i)},\obs^{(i)})$, the loss is $\Loss_{exp}=\sum_{i}\sum_{\states}\exp\{H(\states,\obs^{(i)})-H(\states^{(i)},\obs^{(i)})\}$.
Since for an instance $i$, we are given only the visible part $\visibles^{(i)}$
of $\states^{(i)}$, we formulate the \emph{incomplete} loss as: 
\begin{eqnarray}
\Loss_{inco}(H)=\frac{1}{D}\sum_{i}\left(\sum_{\visibles}\exp\{H(\visibles,\obs^{(i)})-H(\visibles^{(i)},\obs^{(i)})\}\right)^{\beta}\label{exp-loss2}
\end{eqnarray}
Here apart from summing over only the visible component $\visibles$
we also introduce an extra regularisation term $\beta\in(0,1]$ for
numerical stability%
\footnote{We note that a main difference from most of the previous boosting
work is that the number of classes in our cases can be extremely large,
e.g., $|\mathcal{X}|=|S|^{n}$. %
}. In each round $l$ of boosting, the strong learner $H_{l}(\visibles,\obs)$
is updated by adding a `weak-learner' $h_{l}(\visibles,\obs)$ to
the previous $H_{l-1}(\visibles,\obs)$ as $H_{l}(\visibles,\obs)=H_{l-1}(\visibles,\obs)+\alpha_{l}h_{l}(\visibles,\obs)$,
where $\alpha_{l}$ is the weight of each weak learner in the ensemble.
The weak learner and its weight are chosen to minimise the loss in
(\ref{exp-loss2}), i.e. 
\begin{eqnarray}
(h_{l},\alpha_{l}) & = & \arg\min_{h,\alpha}\Loss_{inco}(H_{l-1}+\alpha h)\label{boost-round}
\end{eqnarray}

Since we are interested in the distribution $P(\visibles|\obs)$,
it is sensible to choose the weak learner as $h(\visibles,\obs)=\log P(\visibles|\obs)$.
However, as stated before, if the distribution defined over the general
Markov networks is used, the computation of the weak learner itself
becomes intractable. To this end, we propose the use spanning trees
over graph $\mathcal{G}$ as weak learner serving as an approximation
to the whole network. Thus, each learner is ``weak'' in the sense
that it is an approximation of the true model, but with moderate and
tractable complexity. Let $P_{\tau}(\visibles|\obs)$ be a conditional
distribution of visible state variables $\mathbf{v}$ given the observation
$\mathbf{o}$ with respect to a spanning tree $\tau$, we define the
weak learner to be: 
\begin{eqnarray}
h(\visibles,\obs) & = & \log P_{\tau}(\visibles|\obs)\label{weak-learner}
\end{eqnarray}
This choice also allows incorporation of the missing information since
$h(\visibles,\obs)=\log P_{\tau}(\visibles|\obs)=\log\sum_{\hiddens}P_{\tau}(\visibles,\hiddens|\obs)$.
Thus, the strong learner $H$ is a collection of trees, and we term
our boosting method AdaBoost.MRF (AdaBoosted Markov Random Forests).
Figure~\ref{fig:spanning} shows an example of a simple network and
some spanning trees. We defer the discussion on the choice of tree-based
distributions $P_{\tau}(\visibles|\obs)$ and their relationship to
the model distribution $P(\mathbf{v}|\mathbf{o})$ to later sections.
We shall continue with a derivation on the bound for our formulated
loss in (\ref{exp-loss2}) with the goal to provide a tractable method
to compute this bound.

\subsection{Loss bound using Hölder's inequality}

Although the exponential loss in (\ref{exp-loss2}) is meaningful,
it is unfortunately intractable to compute, let alone minimising it.
In this subsection, we propose to replace the loss by a tractable
upper bound using tree likelihood. Recall that the boosting procedure
is incremental, by substituting (\ref{weak-learner}) into (\ref{exp-loss2})
yields the following expression at step $l$: 
\begin{eqnarray}
\Loss_{inco} & = & \frac{1}{D}\sum_{i}\left(\frac{\sum_{\visibles}\prod_{j=1}^{l}P_{\tau_{j}}(\visibles|\obs^{(i)})^{\alpha_{j}}}{\prod_{j=1}^{l}P_{\tau_{j}}(\visibles^{(i)}|\obs^{(i)})^{\alpha_{j}}}\right)^{\beta}
\end{eqnarray}
The intractability comes from the sum over all visible variables in
the numerator, except for a special case that all selected spanning
trees are the same.

Fortunately, we can get around the summation in the numerator by applying
the Hölder's inequality \cite[Theorem 11]{hardy-et-al52} (see the
appendix for details) to the numerator 
\[
\sum_{\visibles}\prod_{j=1}^{l}P_{\tau_{j}}(\visibles|\obs^{(i)})^{\alpha_{j}}\le\prod_{j=1}^{l}\left[\sum_{\visibles}P_{\tau_{j}}(\visibles|\obs^{(i)})^{\alpha_{j}r_{j}}\right]^{1/r_{j}}
\]
where $\sum_{j=1}^{l}1/r_{j}=1$ and $r_{j}>0$. Under mild assumption
that $\alpha_{j}>0$ and $\alpha_{j}r_{j}=1$ $\forall j$, or $\sum_{j=1}^{l}\alpha_{j}=1$
and note that $1/r_{j}=\alpha_{j},$we obtain: 
\begin{eqnarray}
\Loss_{inco} & \le & \frac{1}{D}\sum_{i}\left(\frac{\prod_{j=1}^{l}\left[\sum_{\visibles}P_{\tau_{j}}(\visibles|\obs^{(i)})\right]{}^{\alpha_{j}}}{\prod_{j=1}^{l}P_{\tau_{j}}(\visibles^{(i)}|\obs^{(i)})^{\alpha_{j}}}\right)^{\beta}\nonumber \\
 & = & \sum_{i}\left(\frac{1}{\prod_{j=1}^{l}P_{\tau_{j}}(\visibles^{(i)}|\obs^{(i)})^{\alpha_{j}}}\right)^{\beta}\nonumber \\
 & = & \frac{1}{D}\sum_{i}\exp\{-\beta\sum_{j=1}^{l}\alpha_{j}\log P_{\tau_{j}}(\visibles^{(i)}|\obs^{(i)})\}
\end{eqnarray}
Let $\Loss_{H}$ be the upper bound on the RHS and given (\ref{weak-learner})
we can further simplify: 
\begin{eqnarray}
\Loss_{H}=\frac{1}{D}\sum_{i}\exp\{-\beta H_{l}(\visibles^{(i)},\obs^{(i)})\}\label{holder-bound}
\end{eqnarray}
since $\sum_{\visibles}P_{\tau_{j}}(\visibles|\obs^{(i)})=1$, $\forall i,j$.
It can be seen that the new bound is tractable to evaluate, and since
is also convex, a global minimum does exist. We thus use the new loss
$\Loss_{H}$ for parameter learning. It can be seen that the domain
of $\Loss_{H}$ is a linear space of functions \cite{Manson-et-al00},
which are $\{h(\visibles,\obs^{(i)})=\log P_{\tau}(\visibles|\obs^{(i)})\}$
in our case.

As we update one weak learner at a step, the requirement $\sum_{j=1}^{l}\alpha_{j}=1$
can be met by defining the following ensemble 
\begin{eqnarray}
H_{l}(\visibles,\obs) & = & (1-\alpha_{l})H_{l-1}(\visibles,\obs)+\alpha_{l}h_{l}(\visibles,\obs)\\
 & = & H_{l-1}(\visibles,\obs)+\alpha_{l}s_{l}(\visibles,\obs)\mbox{ where }\label{additive}\\
s_{l}(\visibles,\obs) & = & h_{l}(\visibles,\obs)-H_{l-1}(\visibles,\obs)\label{new-weak-learner}
\end{eqnarray}
Each previous weak learner's weight is scaled down by a factor of
$1-\alpha_{l}$ as $\alpha'_{j}\leftarrow\alpha_{j}(1-\alpha_{l})$,
for $j=1,...,l-1$, so that $\sum_{j=1}^{l-1}\alpha'_{j}+\alpha_{l}=\sum_{j=1}^{l-1}\alpha_{j}(1-\alpha_{l})+\alpha_{l}=1$
since $\sum_{j=1}^{l-1}\alpha_{j}=1$ .

\subsection{Weak learners, convergence and complexity}

In the previous subsection, we have suggested to use $\Loss_{H}$
as the loss to minimise. Recall that $H(\mathbf{v},\mathbf{o})$ is
a collection of spanning trees (or weak learners), we need to find
the set of trees (and their parameters) that minimise $\Loss_{H}$.
In this subsection, we present an iterative procedure to select trees
that guarantee to reduce the loss. We also provide an analysis of
convergence and time complexity of the procedure.

\subsubsection{Selecting the best tree}

We now show how to carry out the stepwise optimisation in (\ref{boost-round})
with the incomplete loss replaced by the upper bound $\Loss_{H}(H)$
in (\ref{holder-bound}).

The loss $\Loss_{H}(H)$ as a function of $H(\visibles,\obs)$ can
be minimised by moving in the gradient descent direction 
\begin{eqnarray}
\nabla\Loss_{H}(\visibles,\obs^{(i)})=\left\{ \begin{array}{ll}
-\beta\exp\{-\beta H_{l-1}(\visibles^{(i)},\obs^{(i)})\} & \mbox{ if }\visibles=\visibles^{(i)}\\
0 & \mbox{ otherwise}
\end{array}\right.\label{direct-grad2}
\end{eqnarray}
However, as the functional gradient $\nabla\Loss_{H}$ and and the
functional direction $s$ in (\ref{additive}) may not belong to the
same function space, direct optimisation may not apply. In \cite{Manson-et-al00}
the authors propose to find the best $s_{l}$ pointing to the decreasing
direction of $\Loss_{H}$, i.e., 
\begin{eqnarray}
s_{l} & = & \arg\min_{s}\langle\nabla\Loss_{H},s\rangle\label{func-grad}
\end{eqnarray}
subject to 
\begin{equation}
\langle\nabla\Loss_{H},s_{l}\rangle<0\label{eq:stopping}
\end{equation}

The step size $\alpha_{l}$ is determined using a line search or by
setting it to a small constant between $0$ and $1$.

Let $\lambda_{l-1}^{(i)}\propto\exp\{-\beta H_{l-1}(\visibles^{(i)},\obs^{(i)})\}$
be data weights, i.e., $\sum_{i}\lambda_{l-1}^{(i)}=1$. Substituting
(\ref{direct-grad2}) into (\ref{func-grad}), we have 
\begin{eqnarray}
s_{l} & = & \arg\min_{s}\sum_{i}-\lambda_{l-1}^{(i)}s(\visibles^{(i)},\obs^{(i)})
\end{eqnarray}
As $s(\visibles^{(i)},\obs^{(i)})=h(\visibles^{(i)},\obs^{(i)})-H^{l-1}(\visibles^{(i)},\obs^{(i)})$,
minimising with respect to $s(\visibles^{(i)},\obs^{(i)})$ and $h(\visibles^{(i)},\obs^{(i)})$
is equivalent since $H^{l-1}(\visibles^{(i)},\obs^{(i)})$ is a constant.
Recall from (\ref{weak-learner}) that $h(\visibles^{(i)},\obs^{(i)})=\log P_{\tau}(\visibles^{(i)}|\obs^{(i)};\w_{\tau})$,
this minimisation translates to selecting the best tree $t$ and its
parameters $\w_{t}$ as follows: 
\begin{eqnarray}
(t,\w_{t}) & = & \arg\max_{\tau,\w_{\tau}}\sum_{i}\lambda_{l-1}^{(i)}\log P_{\tau}(\visibles^{(i)}|\obs^{(i)};\w_{\tau})\label{max-ll}
\end{eqnarray}

Our final result has a satisfying interpretation: \emph{the functional
gradient descent step tries to solve the maximum re-weighted log-likelihood
problem (\ref{max-ll}) for each tree, and select the best tree with
the largest re-weighted log-likelihood}. As boosting proceeds, some
trees may be more likely to be selected than others, so the accumulated
weights of trees may be different.

As with the standard boosting \cite{schapire99improved}, the data
distribution is iteratively updated as 
\begin{eqnarray}
\lambda_{l}^{(i)} & \propto & \lambda_{l-1}^{(i)}\exp\{-\beta\alpha_{l}s_{l}(\visibles^{(i)},\obs^{(i)})\}
\end{eqnarray}
where $s_{l}$ is the new learner added to the ensemble in (\ref{additive}).
The factor $\beta$ can be used to control the data weights, i.e.,
as $\beta\rightarrow0$, the weights approach the uniform distribution.

Since $\alpha_{l}>0$, the weight increases if $s_{l}=h_{l}-H_{l-1}<0$.
The new interpretation is that \emph{for a given data instance $i$,
if the new weak learner $h_{l}$ is less likely than the average of
previous weak learners $H_{l-1}$, the AdaBoost.MRF will increase
the weight for that data instance}. This is different from the usual
boosting behaviour, where the data weight increases if the strong
learner fails to correctly classify the instance. The AdaBoost.MRF
seems to maximise data likelihood rather than to minimise the training
error, and this is particularly desirable for density estimation.

\subsubsection{Convergence property}

We now provide a formal support for the convergence of the tree selection
procedure in (\ref{max-ll}). The search direction $s$ satisfying
the condition in Equation(~\ref{func-grad}) is called \emph{gradient-related}
to $H^{l}$ \cite[p. 35]{Bertsekas-NLP99}. We have the following
convergence result \cite[Proposition 1.2.3]{Bertsekas-NLP99}.

\label{AdaBoost.MRF-prop:convergence}Given a \emph{Lipschitz continuity}
condition on $\nabla{\Loss}_{H}$, i.e. $\|\nabla{\Loss}_{H}(H)-\nabla{\Loss}_{H}(H')\|\le M\|H-H'\|$,
for some $M>0$, $\forall H,H'\in\mathcal{H}$, where $\mathcal{H}$
is the function space, a gradient-related search direction $s_{l}$,
and a reasonably (positive) small step size $\alpha_{l}$ that satisfies
\begin{eqnarray}
\epsilon\le\alpha_{l}\le(2-\epsilon)\frac{|\langle\nabla{\Loss}_{H}(H_{l-1}),s_{l}\rangle|}{M\|s_{l}\|^{2}}
\end{eqnarray}
where $\epsilon$ is a fixed positive scalar. Then 
\begin{eqnarray}
\lim_{l\rightarrow\infty}H_{l}=\arg\min_{H}{\Loss}_{H}(H)
\end{eqnarray}

The Lipschitz continuity condition can be satisfied in our case because
${\Loss}_{H}$ is twice differentiable, and the Hessian $\nabla^{2}{\Loss}_{H}$
is bounded \cite[p. 48]{Bertsekas-NLP99}. The constant $M$ is hard
to find analytically, so in our implementation, we set the step size
to a small constant $\alpha_{l}=0.05$, and we found it is sufficient
in our experiments. The algorithm terminates when we cannot find any
weak learner $s$ that satisfies the condition in Equation~\ref{func-grad}.

\subsubsection{Complexity}

The running time of AdaBoost.MRF scales linearly in number of trees
$R$, each of which (cf. section \ref{sec:bgr-crf}) takes $\mathcal{O}(2|\vertex||S|^{2})$
inference time. If we only consider limited spanning trees, just enough
to cover the whole network, then $R$ can be quite moderate. For example,
for a fully connected network, we just need $R=|\vertex|$, and in
a grid-like network (e.g., Figure \ref{fig:models}a), $R=2$ is sufficient
(e.g., Figure \ref{fig:two-proc}).

\subsection{Combining the parameters \label{sub:Combining-the-parameters}}

Up to this point, we have successfully estimated the parameters of
individual trees, and thus the strong learner, which may be enough
for classification purposes. However, our ultimate goal is to (approximately)
estimate the parameters of the original network, which is a superimposition
of individual trees. This subsection presents a method for such an
approximate estimation.

Recall that $H(\states,\obs)=\sum_{l}\alpha_{l}h_{l}(\states,\obs)$
and $h_{l}(\states,\obs)=\log P_{\tau_{l}}(\states|\obs)$, and $H(\states,\obs)=\sum_{l}\alpha_{l}\log P_{\tau_{l}}(\states|\obs)$.
Assume that the tree distribution also belongs to the exponential
family in (\ref{expo-dis}) with different parameters $\w_{l}$ and
the same global feature vector $\F(\states,\obs)$. We require that
the parts of the parameters $\w_{l}$, which correspond to cliques
outside the trees, to be zero. We then can rewrite: 
\begin{eqnarray}
H(\states,\obs)=\langle\sum_{l}\alpha_{l}\w_{l},\F(\states,\obs)\rangle-\sum_{l}\alpha_{l}\LogZ_{l}(\obs)
\end{eqnarray}
Thus, the label $\hat{\states}$ returned by the strong learner in
(\ref{max}) becomes 
\[
\hat{\states}=\arg\max_{\mathbf{x}}\langle\sum_{l}\alpha_{l}\w_{l},\F(\states,\obs)\rangle
\]
Obviously, $\hat{\states}$ should also be the MAP assignment of the
model defined by $\states^{{\tiny MAP}}=\arg\max_{\states}P(\states|\obs)=\arg\max\langle\w,\F(\states,\obs)\rangle$,
i.e., $\hat{\states}=\states^{{\tiny MAP}}$. One natural way is to
set $\w$ as the ensemble parameters $\w=\sum_{l}\alpha_{l}\w_{l}$
so that 
\begin{eqnarray}
P(\states|\obs) & \propto & \exp{\langle\sum_{l}\alpha_{l}\w_{l},\F(\states,\obs)\rangle}\nonumber \\
 & \propto & \prod_{l}P_{\tau_{l}}(\states|\obs)^{\alpha_{l}}\\
P(\states|\obs) & = & \frac{\prod_{l}P_{\tau_{l}}(\states|\obs)^{\alpha_{l}}}{\sum_{\states}\prod_{l}P_{\tau_{l}}(\states|\obs)^{\alpha_{l}}}\label{LogOP}
\end{eqnarray}
The combined model turns out to be a Logarithmic Opinion Pool (LogOP)
\cite{heskes98selecting,Pennock-Wellman-UAI99}, a special case of
the more general ensemble framework. Each model $P_{\tau_{l}}(\states|\obs)$
is an expert to provide an estimate of the true distribution $Q(\states|\obs)$.
The aggregator $P(\states|\obs)$ is indeed a minimiser of the weighted
sum of Kullback-Leibler divergences between the $Q(\states|\obs)$
and each $P_{\tau_{l}}(\states|\obs)$ \cite{heskes98selecting} 
\begin{eqnarray}
P(\states|\obs)=\arg\min_{Q(\states|\obs)}\sum_{l}\alpha_{l}\sum_{\states}Q(\states|\obs)\log\frac{Q(\states|\obs)}{P_{\tau_{l}}(\states|\obs)}\label{KL-LogOP}
\end{eqnarray}
A complete set of algorithmic steps for the proposed AdaBoost.MRF
is summarised in Figure~\ref{fig:alg1}.

\subsection{Relations to other works}

In relation to other approaches, the work of \cite{heskes98selecting}
shows that $P(\states|\obs)$ is closer to the true distribution $Q(\states|\obs)$
than the average of all individual experts $P_{\tau_{l}}(\states|\obs)$.
Our boosting algorithm can be seen as an estimator of the weighting
factors $\{\alpha_{l}\}$. \cite{Pennock-Wellman-UAI99} offers an
interesting discussion on the relation between Markov networks and
the LogOP and the properties of desirable aggregators which the LogOP
satisfies. Our method is based on the idea of superimposition, or
\emph{union} of sub-networks, that is, if a node or an edge belongs
to the aggregated network, it must belong to one of the individual
sub-networks. In \cite{Smith-et-alACL05} the authors consider the
combination of different models but they share the same underlying
simple chain structure. Model are trained independently and then combined
using the LogOP. The model weights$\{\alpha_{l}\}$ are then estimated
by maximising the likelihood of the combined models. This approach
is fine as long as the underlying structure is tractable. Another
related idea is the the product-of-experts \cite{Hinton02}, where
all weights are unity. In \cite{Hinton02} sampling is used to overcome
the intractability, which may not converge within a limited time.
By contrast, our method is efficient as it deals directly with trees.

\section{Recognition of Multi-level Activities with Missing Data\label{sec:DCRF-ML}}

\begin{figure}

\begin{centering}
\includegraphics[width=0.45\textwidth]{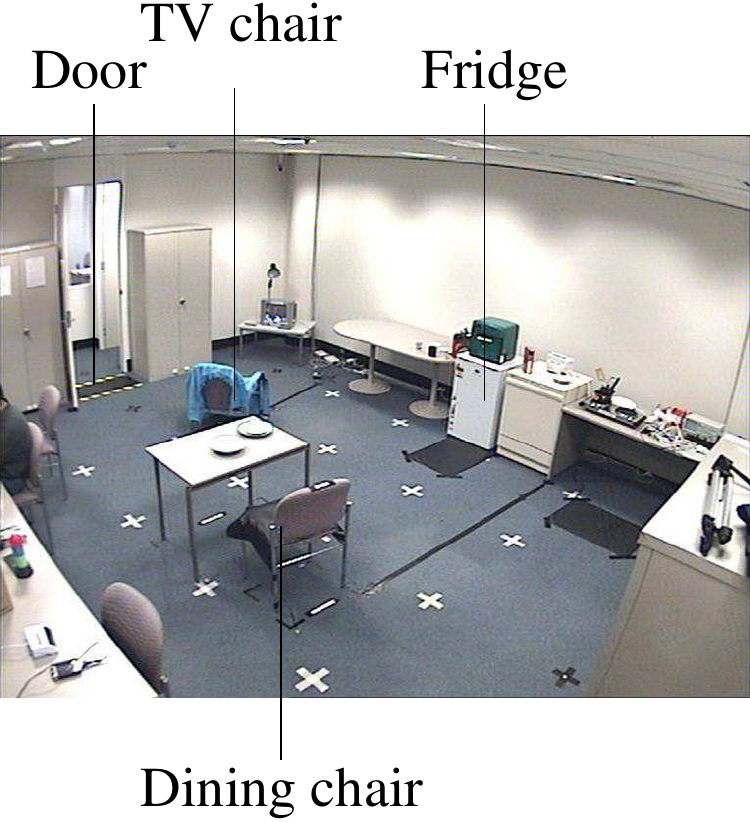}~~~~~\includegraphics[width=0.45\textwidth]{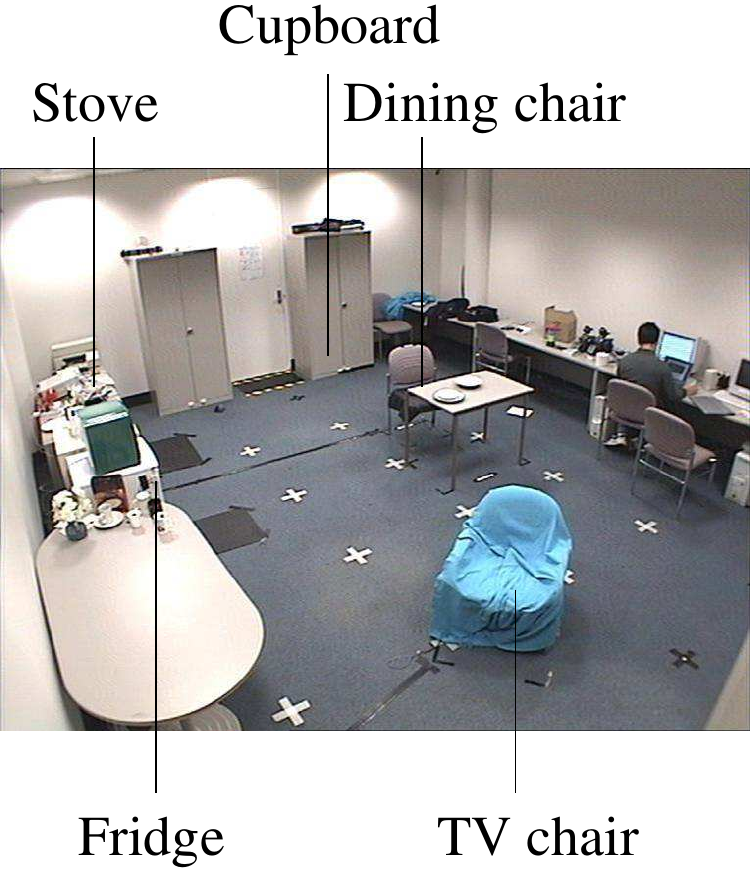}
\par\end{centering}

\caption{The scene viewed from the two cameras~\cite{Nam-et-alCVPR05}. Movement
trajectories of the person are tracked, and the recorded video footage
will be segmented and activities annotated.\label{fig:cam}}

\end{figure}

Our application of interest is to model indoor activities of a person
that are observed through two cameras mounted to the ceilings in a
kitchen, as shown in Figure~\ref{fig:cam}. In this environment,
activities are naturally acted out in a hierarchical manner. We consider
two levels of activity abstraction, which can be modelled using a
two-layer dynamic conditional random field (DCRF) \cite{Sutton-et-alJMLR07}.

\begin{figure}
\begin{centering}
\begin{tabular}{cc}
\includegraphics[width=0.45\textwidth]{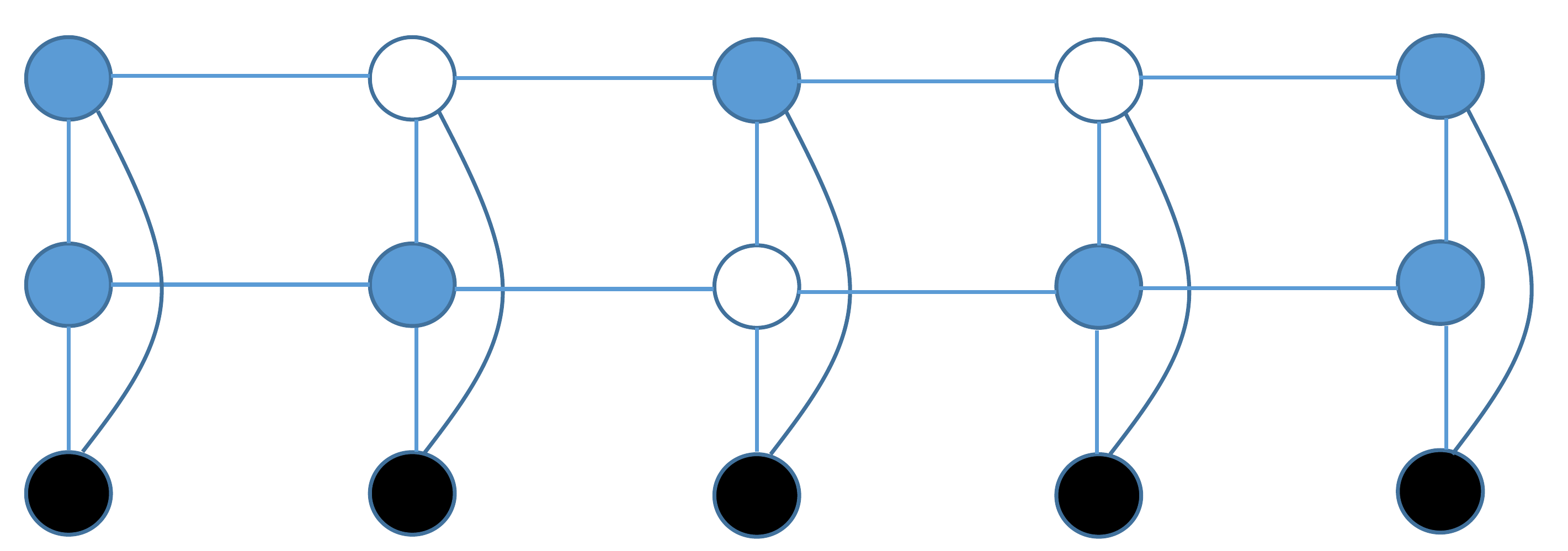} & \includegraphics[width=0.45\textwidth]{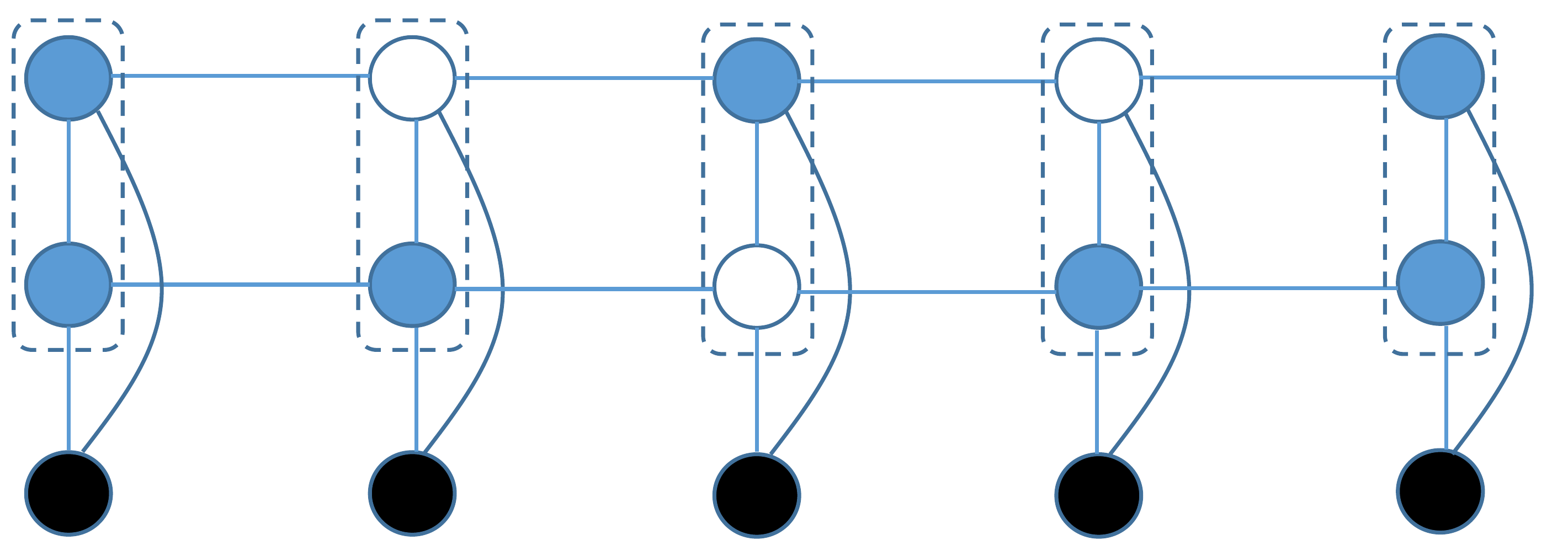}\tabularnewline
(a) ph-DCRF & (b) ph-DCRF\tabularnewline
\end{tabular}
\par\end{centering}

\caption{(a): The partially labelled DCRF (ph-DCRF), and (b): The equivalent
chain CRF. Filled circles and bars are data observations, empty circles
are missing labels, shaded labels are the visible.\label{fig:models}}

\end{figure}

The bottom level presents primitive or atomic activities such as \emph{go-to-cupboard}
or \emph{at-the-fridge}. Higher-order activities are captured at the
higher level such as \emph{having-snack} or \emph{short-meal}. Differing
from the original setting of the DCRF in \cite{Sutton-et-alJMLR07},
we allow some missing labels in our model, and thus we call the model
the partially labelled DCRF (ph-DCRF) (Figure \ref{fig:models}).
We note that although the two-level DCRF is considered in this paper,
the same construction can readily be generalized to model more complex
semantics with richer levels of hierarchy and temporal interactions.

Given the training data, we first learn the parameters and then use
it for annotating and segmenting unseen data. We now describe and
compare some alternatives to the AdaBoost.MRF for parameter learning.

For the original ph-DCRF, exact estimation of marginals can be carried
out by collapsing all the states at the current time into a mega-state
(see Figure~\ref{fig:models}b) and performing a \emph{forward-backward
}procedure, which is infeasible for deep models. Approximate inference
using the BP and WJW ~\cite{Wainwright-et-alIEEEIT05} methods has
the complexity of $\mathcal{O}(2I|\edges||S|^{2})$, where $I$ is
the number of message passing rounds, $|\edges|$ is the number of
edges in the network, and $S$ is the state size per node. However,
the number of rounds $I$ until convergence if it does is not known
analytically, and there has not been any theoretical estimate of it
yet.

In our AdaBoost.MRF, inference in the trees takes $\mathcal{O}(2|\vertex||S|^{2})$
time, where $|\vertex|$ is the number of nodes in the network. Thus,
for $D$ data instances, and $R$ trees, the AdaBoost.MRF costs $\mathcal{O}(4DR|\vertex||S|^{2})$
in total time for each gradient evaluation since we need to take both
$\LogZ(\visibles,\obs)$ and $\LogZ(\obs)$ into account. Similarly,
the BP and WJW-based MLE requires $\mathcal{O}(4DI|\edges||S|^{2})$
time. As for fully connected networks, $|\edges|$=$\frac{1}{2}|\vertex|(|\vertex|+1)$
while for the grid DCRFs, $|\edges|\approx2|\vertex|$, if we take
only $R=|\vertex|$ trees for the former case, and $R=2$ for the
latter case, the total complexity per gradient evaluation of the BP
and WJW-based MLE and the AdaBoost.MRF will be similar up to a constant
$I$. We summarise the complexities in Table \ref{tab:complexity}.

\begin{table}[htb]
\centering{}%
\begin{tabular}{|c|c|}
\hline 
BP/WJW & AdaBoost.MRF\tabularnewline
\hline 
$\mathcal{O}(4DI|\edges||S|^{2})$  & $\mathcal{O}(4DR|\vertex||S|^{2})$ \tabularnewline
\hline 
\end{tabular}\caption{Complexity per gradient evaluation compared with existing message-passing
methods.\label{tab:complexity} }
\end{table}

\section{Experimental Results\label{sec:exp}}

The dataset used in this experiment was collected in our previous
work \cite{Nam-et-alCVPR05} using a system shown in Figure~\ref{fig:cam}.
We captured 45 video sequences for training and 45 sequences for testing.
The observations are sequences of noisy coordinates of the actor walking
in scene acquired using a background subtraction tracking algorithm.
We consider 3 complex activities (states) at the top level: \emph{short-meal}
(1), \emph{have-snack} (2), \emph{normal-meal (3),} and twelve primitive
activities at the bottom level, which are summarised in Table~\ref{tab:pri_beh}.

Each complex activity is comprised of some primitive activities, and
states at each level can freely transit to each other but generally
we do not have this knowledge at hand for our experiment.

For evaluating the aspect of missing labels, we randomly provide half
the labels for each level during training. For testing, the MAP assignments
resulted from Pearl's loopy max-product algorithm are compared against
the ground-truth.

\begin{table}[htb]
\centering{}%
\begin{tabular}{|c|l|c|l|}
\hline 
{\small No.} & {\small Activity }  & {\small No.}  & {\small Activity}\tabularnewline
\hline 
{\small 1}  & {\small Door$\to$Cupboard}  & {\small 7}  & {\small Fridge$\to$TV chair}\tabularnewline
{\small 2}  & {\small Cupboard$\to$Fridge }  & {\small 8}  & {\small TV chair$\to$Door}\tabularnewline
{\small 3}  & {\small Fridge$\to$Dining chair}  & {\small 9}  & {\small Fridge$\to$Stove}\tabularnewline
{\small 4}  & {\small Dining chair$\to$Door}  & {\small 10}  & {\small Stove$\to$Dining chair}\tabularnewline
{\small 5}  & {\small Door$\to$TV chair }  & {\small 11}  & {\small Fridge$\to$Door}\tabularnewline
{\small 6}  & {\small TV chair$\to$Cupboard}  & {\small 12}  & {\small Dining chair$\to$Fridge}\tabularnewline
\hline 
\end{tabular}\caption{Primitive activities used in the experiment.\label{tab:pri_beh}}
\end{table}

\subsection{Feature Extraction}

With the data described above, the input to the DCRFs is simply sequences
of coordinates. At each time slice $\gamma$, we extract a vector
of five elements from the observation sequence $\mathbf{g}(\obs,\gamma)=\{g_{m}(\obs,\gamma)\}_{m=1}^{5}$
where each $g_{m}(\obs,\gamma)$ corresponds to the $(X,Y)$ coordinates,
the $X$ and $Y$ velocities, and the speed respectively. To fully
specify the model, we consider three types of feature functions for
the potentials of the network: (a) \emph{data-association} corresponding
to node potentials, (b) \emph{temporal-relation} corresponding to
state transition potentials at the same level, and (c) \emph{cross-semantic-relation}
corresponding to parent-child potentials across different levels.

Let $x_{\gamma}^{2}$ denotes the state variable at the level 2 (the
bottom) and time $\gamma$. For the first feature set, we define the
data-association features at the bottom level as:

\begin{align}
\f_{i,m,\epsilon}(x_{\gamma}^{2},\obs):=g_{m}(\obs,\gamma+\epsilon)\mathbb{I}\left\{ x_{\gamma}^{2}=i\right\} \label{data-ass-feature}
\end{align}
where $\epsilon=-s_{1},...,0,...,s_{2}$ is the amount of look-ahead
or look-back, for some positive integers $s_{1}$, $s_{2}$, $x_{\gamma}^{2}=1,2,...,12$
is the state (at level 2). We choose $s_{1}=s_{2}=2$ for reasonable
computation, so that the current primitive activity $x_{\gamma}^{2}$
is correlated with five surrounding observation features $g_{m}(\obs,\gamma)$.
At the top level, however, instant information such as velocities
offer limited help since the complex activities often span long periods.
Instead of using the real coordinates $(X,Y)$ for data association,
we quantize them into 24 squares in the room. We also use much larger
windows with $s_{1}=s_{2}=20$. To avoid computational overhead, we
take $\epsilon=-s_{1},-s_{1}+5,...,s_{2}-5,s_{2}$.

The second and third feature sets consist of simple indicator functions

\begin{align*}
\f_{l_{1},l_{2}}(x_{\gamma-1}^{d},x_{\gamma}^{d}):=\mathbb{I}\left\{ x_{\gamma-1}^{d}=l_{1}\right\} \mathbb{I}\left\{ x_{\gamma}^{d}=l_{2}\right\} 
\end{align*}
for the second set, and

\begin{align*}
\f_{l_{pa},l_{ch}}(x_{\gamma}^{d-1},x_{\gamma}^{d}):=\mathbb{I}\left\{ x_{\gamma}^{d-1}=l_{pa}\right\} \mathbb{I}\left\{ x_{\gamma}^{d}=l_{ch}\right\} 
\end{align*}
for the third set, where $d=1,2$ is the depth level.

\subsection{Spanning trees for AdaBoost.MRF \label{sec:union}}

\begin{figure}
\begin{centering}
\begin{tabular}{cc}
\includegraphics[width=0.45\textwidth]{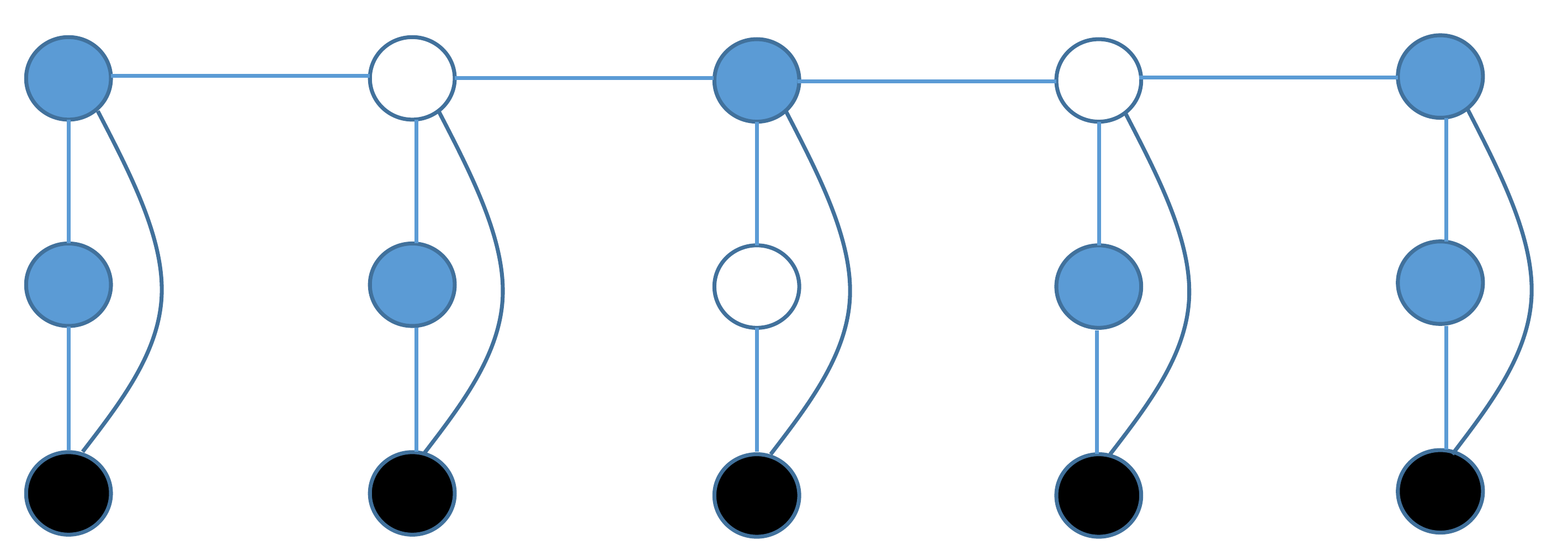} & \includegraphics[width=0.45\textwidth]{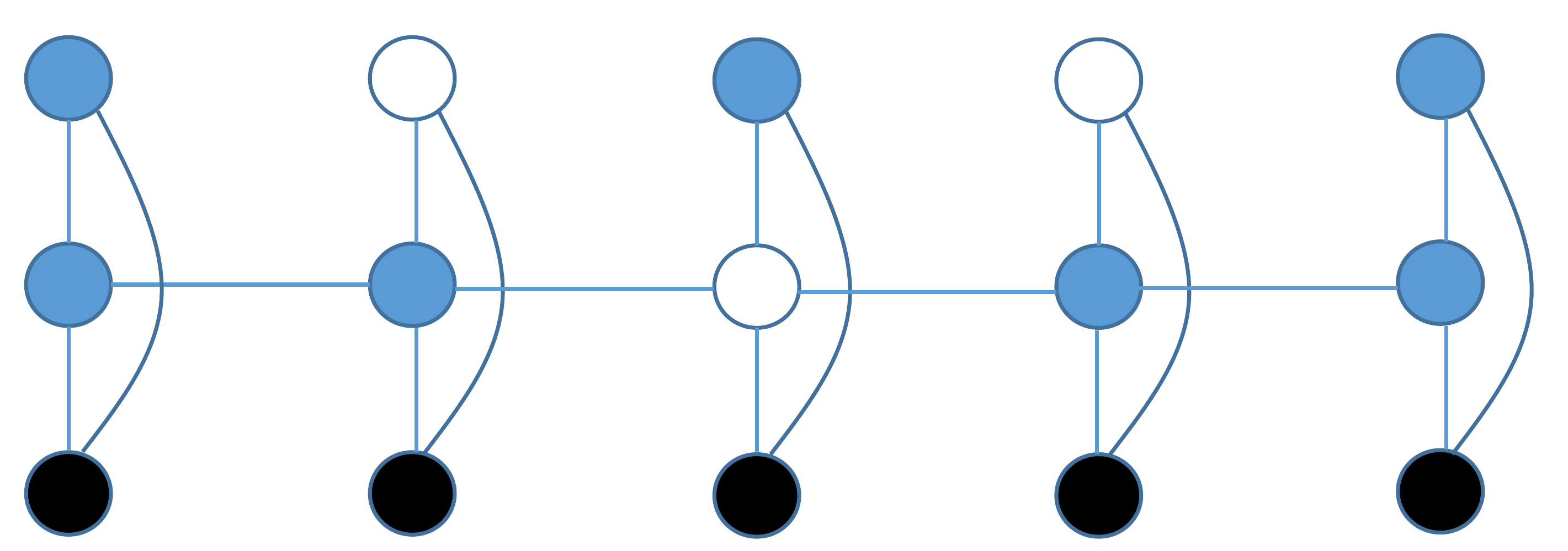}\tabularnewline
(a) Top process & (b) Bottom process\tabularnewline
\end{tabular}
\par\end{centering}

\caption{Two process view of the DCRF in activity modeling: (a) the complex
activity, and (b) the primitive.\label{fig:two-proc}}

\end{figure}

The AdaBoost.MRF algorithm described in Figure~\ref{fig:alg1} requires
the specification of a set of spanning trees which will be used as
the weak classifiers. Given the grid structure considered in this
experiment, there are many spanning trees that can be extracted. However,
since the nature of our problem is about temporal regularities where
the slice structure is repeated over time, it is natural to decompose
the network into trees in a such a way that the structural repetition
is maintained. With this hint, there are two most noticeable trees
that stand out as shown in Figure \ref{fig:two-proc}, which roughly
corresponds the top and bottom chains respectively.

\begin{figure}[H]
\begin{centering}
\includegraphics[width=0.09\linewidth]{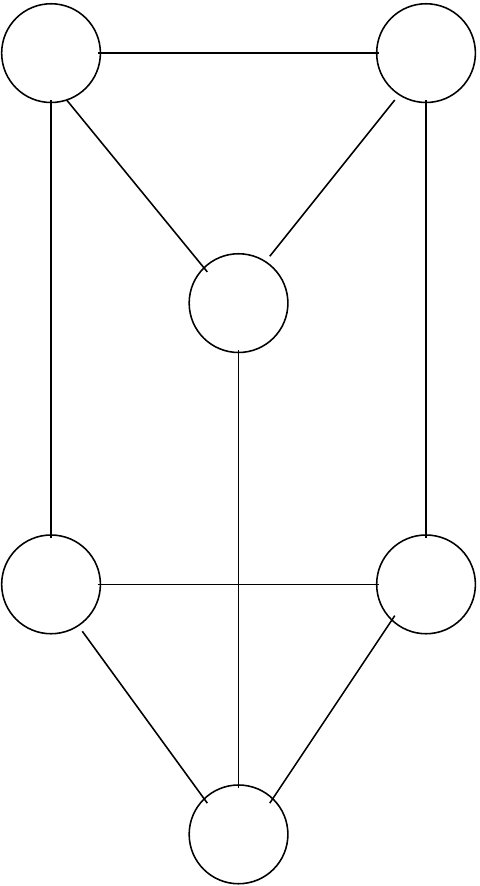}~~~~~~~\includegraphics[width=0.1\linewidth]{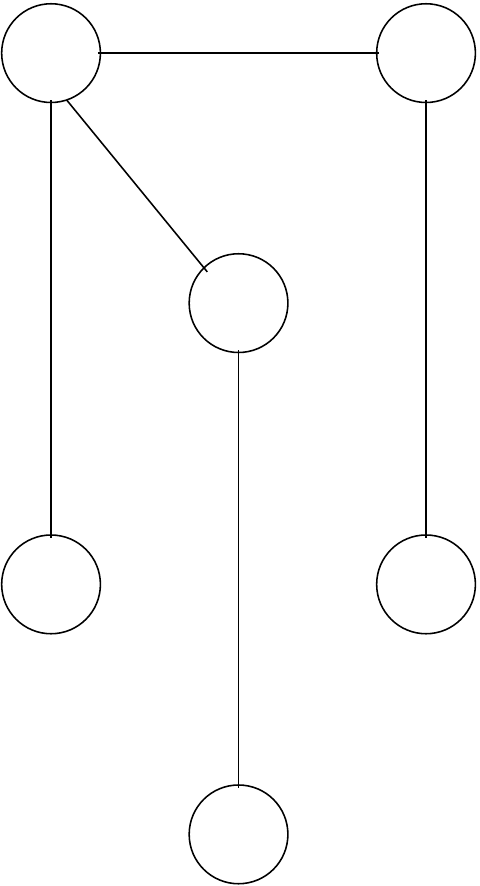}~~~~~~~\includegraphics[width=0.1\linewidth]{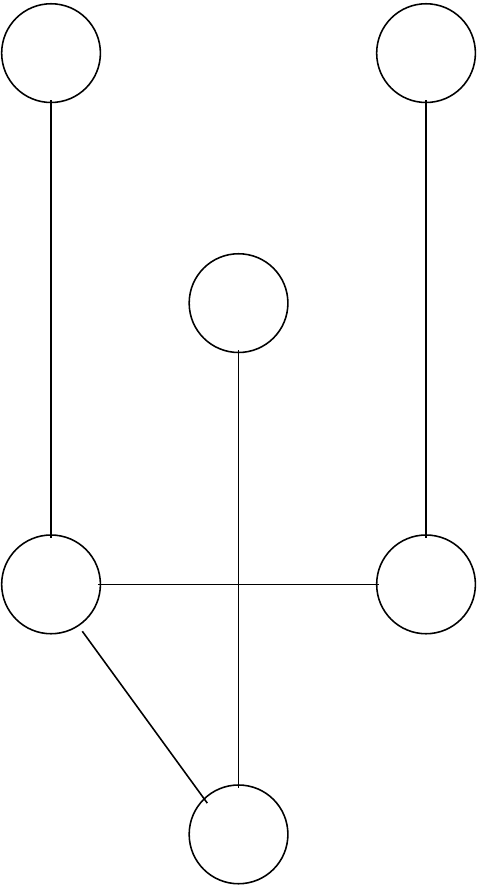}~~~~~~~\includegraphics[width=0.1\linewidth]{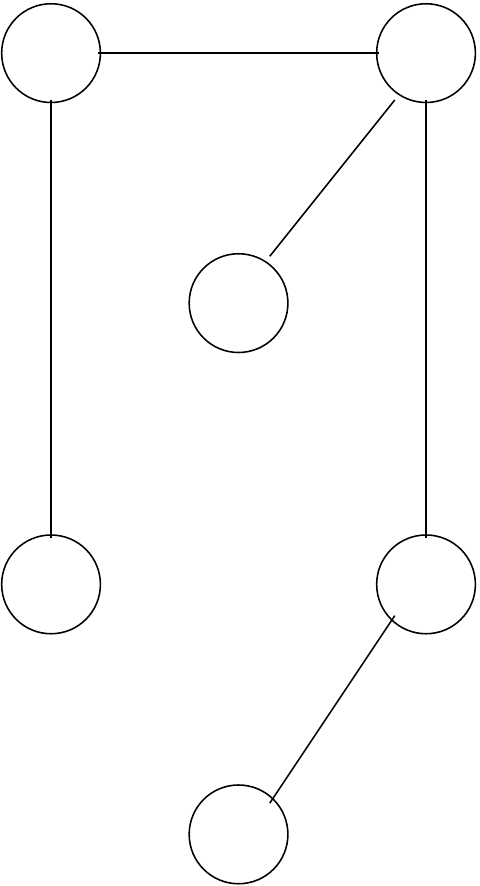}~~~~~~~\includegraphics[width=0.1\linewidth]{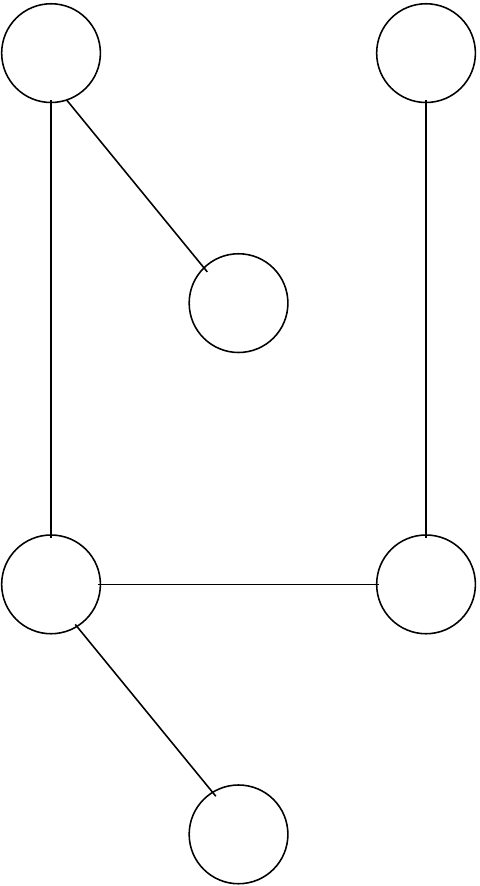} 
\par\end{centering}

\caption{Two-slice structure of a DCRF (left-most), and some spanning trees
(the rest).\label{fig:trees}}
\end{figure}

With the same method, the number of trees for dynamic models which
respect the Markov assumption is reduced drastically. If we impose
further restrictions that each state can only interact with the level
right above and right below it, then the number of trees can be manageable
(e.g. see Figure \ref{fig:trees} for another example).

\subsection{Segmentation and annotation results}

For comparing with the AdaBoost.MRF for the DCRFs, we implement MLE
learning methods based on BP, WJW and exact inference. We also evaluate
the effectiveness of the DCRFs against the Layered HMMs (LHMMs) \cite{oliver2004lrl},
where the output of the bottom HMM is used as the input for the top
HMM. Since, it is difficult to encode rich feature information in
the LHMMs without producing very large state space, we limit the LHMMs
features to be the discretised positions and the differences between
current position and the previous and next ones. Our new implementation
of LHMMs differs from the original in \cite{oliver2004lrl} for each
HMM has been extended to handle the partially observed states. To
test whether adding more layers can improve the performance of the
model, we run a simple \emph{Flat-CRF} on the data at the lower level.
All learning algorithms are initialised uniformly. For segmentation
purposes, we report the macro-averaged $F_{1}$ scores on a per-label
basis.

\begin{figure}[H]
\begin{centering}
\includegraphics[width=0.7\textwidth]{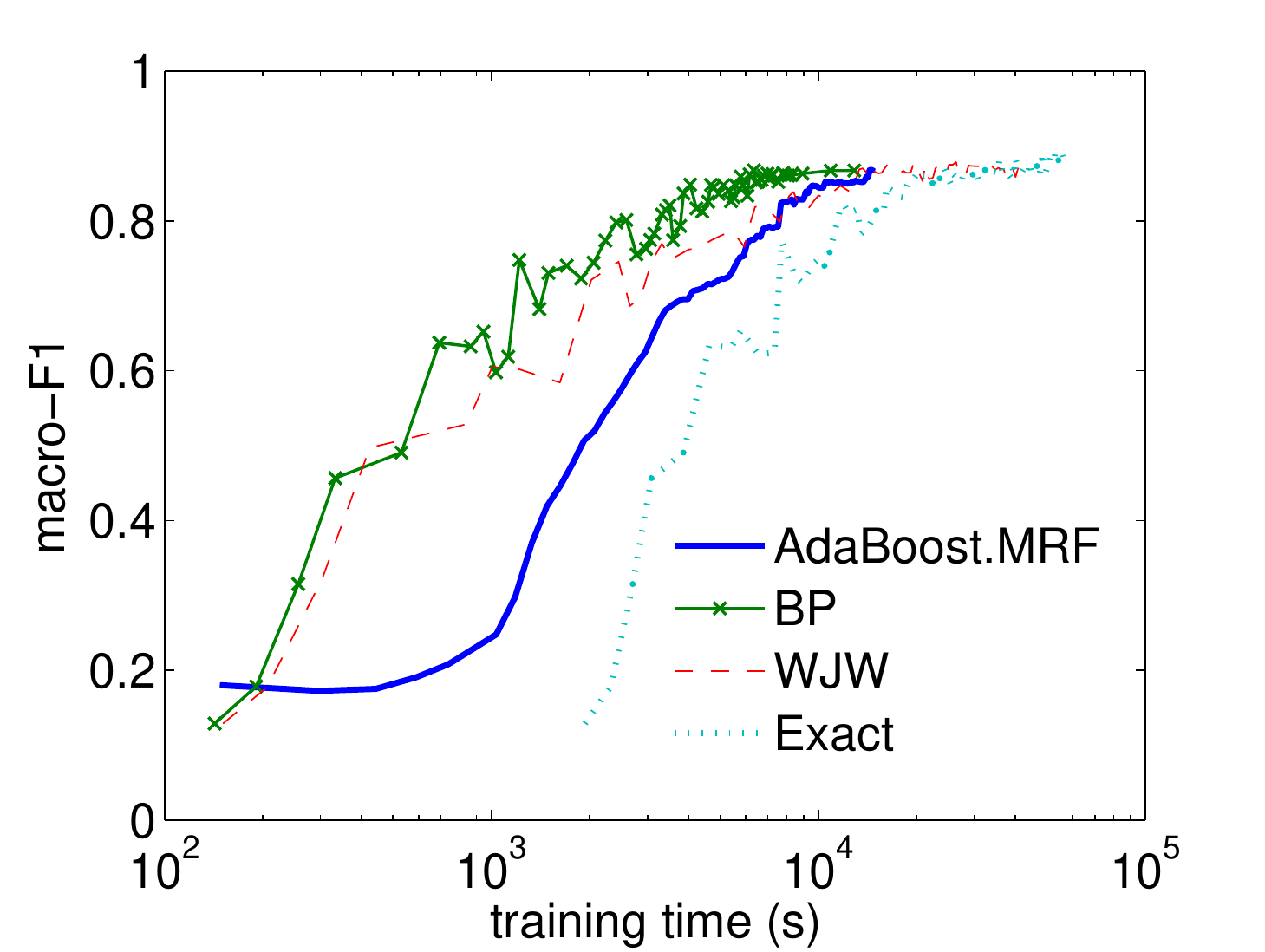}
\par\end{centering}

\caption{Macro-averaged $F_{1}$ scores at the bottom layer vs. training time.\label{fig:train-time}}
\end{figure}

For parameter optimisation of the (re-weighted) log-likelihood, initially
we used the limited memory quasi-Newton method (L-BFGS) as suggested
in the CRF literature but it seems to be slower and it converges prematurely
to poor solutions for the BP and the exact inference. The conjugate-gradient
(CG) method works better in our experiments. For the Markov forests,
we run for only two iterations of CG per boosting round with the initial
parameters from the previously learned ones since we only need to
meet the condition (\ref{func-grad}). The WJW inference loop is stopped
if the messages have converged at the rate of $10^{-4}$ or after
100 rounds. It appears that the final performance of BP is sensitive
to the choice of convergence rates, while it is fairly stable for
the WJW. For example, the $F_{1}$ scores at the bottom level for
BP are $0.84,0.87$ and $0.82$ corresponding to the rates of $10^{-3},10^{-4}$
and $10^{-5}$, respectively. Below we report only the case of $10^{-4}$,
which appears to be the best both in terms of accuracy and speed.
Learning algorithms for the DCRFs are stopped after 100 iterations
if they have not converged at the rate of $10^{-5}$.

The performance of the AdaBoost.MRF and its alternatives is reported
in Figure~\ref{fig:train-time} and Table~\ref{all-results}, respectively.
Overall, after enough training time, the AdaBoost.MRF performs comparably
with the MLE methods based on BP and WJW. The exact inference MLE
method gives slightly better result as expected but at the cost of
much slower training time. However, it should be stressed that inference
in our AdaBoost.MRF always converges, while it is not guaranteed in
the BP and WJW and it is generally intractable in the exact method.
The complexity per evaluation of the log-likelihood gradient is known
and fixed for the AdaBoost.MRF, while for the BP and the WJW, it is
generally dependent on the convergence criteria and how much the distribution
is different from uniform (see Table~\ref{tab:complexity}).

Table~\ref{all-results} also shows that the choice of discriminative
model over the generative model in our activity recognition problem
is justified. The LHMMs are worse than both the flat-CRFs at the bottom
layer and the DCRFs at the top layer. Furthermore, the DCRFs variants
are more consistently accurate than the flat CRFs. The result is consistent
with that in \cite{Sutton-et-alJMLR07}. This can be explained by
the fact that more information is encoded in the DCRFs.

\begin{table}
\centering{}%
\begin{tabular}{|l||c|c|}
\hline 
Algorithm  & Top-layer  & Bottom-layer\tabularnewline
\hline 
\hline 
AdaBoost.MRF  & 0.98  & 0.87\tabularnewline
\hline 
BP  & 0.99  & 0.87\tabularnewline
\hline 
WJW  & 0.98  & 0.87\tabularnewline
\hline 
Exact  & 0.98  & 0.88\tabularnewline
\hline 
LHMM  & 0.88  & 0.67\tabularnewline
\hline 
Flat-CRFs  & -  & 0.78\tabularnewline
\hline 
\end{tabular}\caption{Macro-averaged $F_{1}$ scores for top and bottom layers. \label{all-results} }
\end{table}

\begin{figure}[H]
\begin{centering}
\includegraphics[scale=0.8]{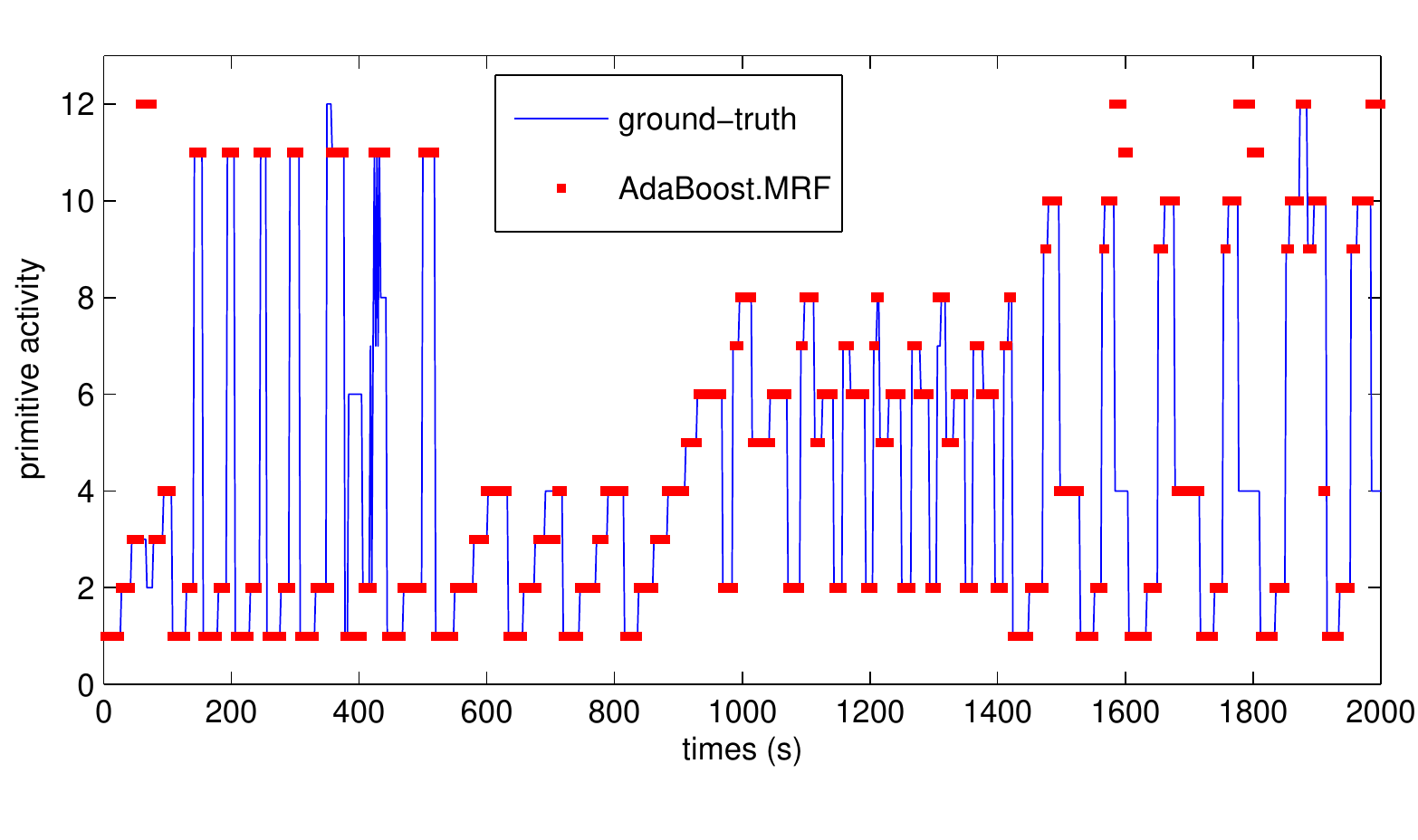}
\par\end{centering}

\caption{The segmentation compared with the ground-truth at the bottom level.\label{fig:seg-levels}}
\end{figure}

Figure \ref{fig:seg-levels} shows the AdaBoost.MRF segmentation details
of 22 randomly selected sequences which are concatenated together.

\section{Discussion\label{sec:discuss}}

In this section we would like to further discuss the implications
of the proposed AdaBoost.MRF as a guided search for estimation in
maximum likelihood setting and as a method for feature selection.

\subsection{AdaBoost.MRF as guided search for maximum likelihood estimation\label{sec:guide-MLE}}

As we rely on the boosting capacity to boost very weak learners to
a strong one, we do not need to reach the maximum of the weighted
log-likelihood in each round. We can simply run a few training iterations
and take the partial results as long as the condition in (\ref{func-grad})
is met. To speedup the learning, we can initialise the parameters
for each weak learner to the previously learned values. This procedure
has an interesting interpretation for tree-structured graphs. As we
do not have to select the best spanning trees anymore, the algorithm
is simply to optimise the re-weighted log-likelihood in a stage-wise
manner. We argue that this approach can be attractive because more
information from the data distribution can be used to guide the maximum
likelihood estimation (MLE), and it can create more diverse weak classifiers.

For non-tree graphs, the guided MLE can be derived by assuming that
the tree mixing coefficients $\alpha^{l}$ are known in advance. Recall
that we have assumed that parameters are not shared among trees. To
see how we can relax this assumption by allowing trees to share some
common parameters, let $\w$ be the joint parameters of all the trees,
we rewrite the loss in (\ref{holder-bound}) as:

\begin{align}
\Loss_{H}(\w)=\frac{1}{D}\sum_{i}\exp\{-\beta\alpha^{l}\log P^{l}(\visibles^{(i)}|\obs^{(i)};\w)\}\label{eq:holder-bound}
\end{align}

Taking derivative of $\Loss_{H}(\w)$ with respect to $\w$ yields:

\begin{align}
\nabla\Loss_{H}(\w)\propto-\sum_{i}\lambda^{(i)}\sum_{t}\alpha^{l}\nabla\log P^{l}(\visibles^{(i)}|\obs^{(i)};\w)\label{eq:holder-bound2-grad}
\end{align}
Thus we have arrived a parallel version of the AdaBoost.MRF in that
all trees are updated at the same time. This version, however, loses
the tree-selection property.

\subsection{AdaBoost.MRF as embedded method for feature selection}

It should be noted that in our formulation of AdaBoost.MRF, we do
not require the unique existence of one distribution per tree. Instead,
we have to freedom to choose as many distributions as we wish, provided
that these distributions can be expressed in the log-linear form (cf.
section~\ref{sec:bgr-crf}). Thus, given a feature pool, we can create
a set of distributions, each of which incorporates only a subset of
features. As the AdaBoost.MRF proceeds, only one particular tree-based
distribution is picked at a time, thus implicitly performing the feature
selection capacity. It is reasonable to expect that bad feature combination
will result in a poor likelihood, and thus will not be selected in
Equation(~\ref{max-ll}).

\section{Conclusion\label{sec:con}}

We have presented a novel method for using boosting in parameter estimation
of the general Markov networks with partial labels. The algorithm
AdaBoost.MRF offers an efficient way to tackle the intractability
of the maximum likelihood method by breaking the model into tractable
trees and combining them to recover the original networks. We apply
the algorithm to the new problem of multi-level activity recognition
and segmentation using the recently proposed DCRFs.

We would like to stress, however, that our AdaBoost.MRF is not limited
to only DCRFs but can be readily applied to any arbitrary CRFs. In
addition, not only it can discriminatively approximately estimate
the conditional distributions $P(\states|\obs)$, but also it can
generatively learn the \emph{joint} distributions $P(\states,\obs)$.

Furthermore, in our experiments, it appears that the AdaBoost.MRF
exhibits a structure learning behaviour since it may selectively pick
some trees more frequently than others, giving higher weights to those
trees. An important issue we have left unanswered is how to automatically
select the optimal tree at each round without knowing the set of trees
in advance. We plan to investigate these aspects and the use of AdaBoost.MRF
in a wider range of applications.

\appendix

\section{General Hölder's Inequality\label{sec:holder-proof}}

Let us start with the elementary Hölder's inequalities \cite[Theorem 13]{hardy-et-al52}.
For $r>1,r'>1,a_{i}\ge0,b_{i}\ge0$ and $1/r+1/r'=1$, the following
inequality holds:

\begin{align}
\sum_{i=1}^{n}a_{i}b_{i}\le(\sum_{i=1}^{n}a_{i}^{r})^{1/r}(\sum_{i=1}^{n}b_{i}^{r'})^{1/r'}\label{basic-holder}
\end{align}

The sign of equality hold iff $a_{i}^{r}=\alpha b_{i}^{r'},\forall i$,
for some scalar $\alpha$. The case $\alpha=0$ is trivial, thus we
do not consider here. The Cauchy's inequality is a special case if
$r=r'=2$.

By induction, we can obtain the following extension to this basic
inequality \cite[Theorem 11]{hardy-et-al52}. If $a_{ij}\ge0$, for
$i=1,2,..,n$ and $j=1,2,..,m$ , and if $r_{j}>1$ with $\sum_{j=1}^{m}1/r_{j}=1$,
denoting the vector $A_{j}=(a_{1j},a_{2j},\ldots,a_{nj})$, then

\begin{align}
\sum_{i=1}^{n}\prod_{j=1}^{m}a_{ij}\le\prod_{j=1}^{m}(\sum_{i=1}^{n}a_{ij}^{r_{j}})^{1/r_{j}}\label{general-holder}
\end{align}
and the sign of equality holding iff $A_{j}=\alpha_{jk}A_{k}$ for
some scalars $\alpha_{jk}$ and $j\neq k$. In other words, the equality
sign holds iff all vectors $A_{j}$(s) are proportional.

Let us proceed by induction to prove the `inequality' part.

(i) For $j=1$, (\ref{general-holder}) holds trivially.

(ii) Assume (\ref{general-holder}) holds for any $1\leq j\le l$,
we will prove that it also holds for $j=l+1$. Using the basic Hölder
inequality (\ref{basic-holder}) for $r>1;r'>1;1/r+1/r'=1$, we have:

\begin{align}
\sum_{i}\prod_{j=1}^{l+1}a_{i,j}=\sum_{i}a_{i,l+1}\prod_{j=1}^{l}a_{i,j}\le & \left(\sum_{i}a_{i,l+1}^{r}\right)^{1/r}\left(\sum_{i}\left[\prod_{j=1}^{l}a_{i,j}\right]^{r'}\right)^{1/r'}\\
 & =\left(\sum_{i}a_{i,l+1}^{r}\right)^{1/r}\left(\sum_{i}\prod_{j=1}^{l}a_{i,j}^{r'}\right)^{1/r'}\label{holder-proof1}
\end{align}
Let $\beta_{i,j}=a_{i,j}^{r'}$, by applying the inductive assumption
that (\ref{general-holder}) holds for $j\le l$ to the second factor
in the RHS of (\ref{holder-proof1}) yields:

\begin{align}
\left(\sum_{i}\prod_{j=1}^{l}\beta_{i,j}\right)^{1/r'} & \le\left(\prod_{j=1}^{l}\left[\sum_{i}\beta_{i,j}^{r'_{j}}\right]^{1/r'_{j}}\right)^{1/r'}\nonumber \\
 & =\prod_{j=1}^{l}\left[\sum_{i}\beta_{i,j}^{r'_{j}}\right]^{1/r'_{j}r'}=\prod_{j=1}^{l}\left[\sum_{i}a_{i,j}^{r'_{j}r'}\right]^{1/r'_{j}r'}\label{holder-proof2}
\end{align}
where $\sum_{j=1}^{l}1/r'_{j}=1$. Substituting (\ref{holder-proof2})
back into (\ref{holder-proof1}), we have:

\begin{align}
\sum_{i}\prod_{j=1}^{l+1}a_{i,j} & \le\left(\sum_{i}a_{i,l+1}^{r}\right)^{1/r}\prod_{j=1}^{l}\left[\sum_{i}a_{i,j}^{r'_{j}r'}\right]^{1/r'_{j}r'}\label{holder-proof3}
\end{align}

As $\sum_{j=1}^{l}1/r'_{j}=1$, we then have $1/r+\sum_{j=1}^{l}1/r'_{j}r'=1/r+1/r'=1$.
Now we change the notation as $r'_{l+1}\leftarrow r$ and $r'_{j}\leftarrow r'_{j}r'$,
we have $\sum_{j=1}^{l+1}1/r'_{j}=1$. Thus (\ref{holder-proof3})
becomes

\begin{align}
\sum_{i}\prod_{j=1}^{l+1}a_{i,j} & \le\prod_{j=1}^{l+1}\left[\sum_{i}a_{i,j}^{r'_{j}}\right]^{1/r'_{j}}\label{holder-proof4}
\end{align}

This means that the inequality (\ref{general-holder}) holds for $j=l+1$.
By the induction principle, the inequality (\ref{general-holder})
holds for all $j\ge1$. This completes the proof.\hfill{} $\diamondsuit$

\section{Pseudo-code for AdaBoost.MRF}

\begin{figure}[H]
\begin{centering}
\begin{tabular}{l}
\hline 
\textbf{Input}: $i=1,...,D$ data pairs, graphs $\{\graph^{(i)}=(\vertex^{(i)},\edges^{(i)})\}$\tabularnewline
$\quad$$\quad$$\quad$$\quad$$\quad$$\quad$and the regularisation
term $\beta\in(0,1]$\tabularnewline
\textbf{Output}: learned parameter vector $\w$\tabularnewline
\textbf{Begin} \tabularnewline
$\quad$$\quad$Select spanning trees for each data instance \tabularnewline
$\quad$$\quad$ Initialise $\{\lambda_{0}^{(i)}=\frac{1}{D}\}$,
and $\alpha_{1}=1$ \tabularnewline
$\quad$$\quad$\textbf{For} each boosting round $l=1,2,\dots$ \tabularnewline
$\quad$$\quad$$\quad$$\quad$ Train all trees given weighted data
$\{\lambda_{l-1}^{(i)}\}_{i=1}^{D}$\tabularnewline
$\quad$$\quad$$\quad$$\quad$\em /{*}Select the best tree distribution{*}/\em \tabularnewline
$\quad$$\quad$$\quad$$\quad$$(t,\w_{t})=\arg\max_{\tau,\w_{\tau}}\sum_{i}\lambda_{l-1}^{(i)}\log P_{\tau}(\visibles^{(i)}|\obs^{(i)};\w_{\tau})$ \tabularnewline
$\quad$$\quad$$\quad$$\quad$$h_{l}=\log P_{t}(\visibles|\obs;\w_{t})$ \tabularnewline
$\quad$$\quad$$\quad$$\quad$$s_{l}=h^{l}-H^{l-1}$ \tabularnewline
$\quad$$\quad$$\quad$$\quad$\textbf{If} $\sum_{i}\lambda_{l-1}^{(i)}s_{l}(\visibles^{(i)}|\obs^{(i)})\le0$
\textbf{Then} go to Output \tabularnewline
$\quad$$\quad$$\quad$$\quad$\textbf{If} $l>1$ \textbf{Then} select
the step size $0<\alpha_{l}<1$ \tabularnewline
$\quad$$\quad$$\quad$$\quad$\em /{*}Update the strong learner{*}/\em \tabularnewline
$\quad$$\quad$$\quad$$\quad$$H_{l}=(1-\alpha_{l})H_{l-1}+\alpha_{l}h_{l}$ \tabularnewline
$\quad$$\quad$$\quad$$\quad$\em /{*}Scale down the previous learner
weights{*}/\em \tabularnewline
$\quad$$\quad$$\quad$$\quad$$\alpha_{j}\leftarrow\alpha_{j}(1-\alpha_{l})$,
for $j=1,...,l-1$\tabularnewline
$\quad$$\quad$$\quad$$\quad$\em /{*}Update the data weight{*}/\em \tabularnewline
$\quad$$\quad$$\quad$$\quad$$\lambda_{l}^{(i)}\leftarrow\lambda_{l-1}^{(i)}\exp\{-\beta\alpha_{l}s_{l}(\visibles^{(i)}|\obs^{(i)})\}$\tabularnewline
$\quad$$\quad$$\quad$$\quad$ $\lambda_{l}^{(i)}\leftarrow\frac{\lambda_{l}^{(i)}}{\sum_{i=1}^{D}\lambda_{l}^{(i)}}$ \tabularnewline
$\quad$$\quad$\textbf{End} \tabularnewline
$\quad$$\quad$Output $\w=\sum_{l}\alpha_{l}\w_{l}$ \tabularnewline
\textbf{End} \tabularnewline
\hline 
\end{tabular}
\par\end{centering}

\caption{AdaBoost.MRF - AdaBoosted Markov Random Forests.\label{fig:alg1}}
\end{figure}

\end{document}